\documentclass[11pt]{article}

\usepackage[letterpaper]{geometry}

\usepackage{natbib}
\usepackage{placeins} %float barriers

\usepackage{url}
\usepackage{amsmath}
\usepackage{amssymb}
\usepackage{graphicx,soul,color}
\usepackage{booktabs}

\newcommand{\bftab}{\fontseries{b}\selectfont}

\author{
  George E.~Dahl\\
  \texttt{gdahl@cs.toronto.edu}
  \and
  Navdeep Jaitly\\
  \texttt{ndjaitly@cs.toronto.edu}
  \and
  Ruslan Salakhutdinov\\
  \texttt{rsalakhu@cs.toronto.edu}\\
\\
\normalsize{Department of Computer Science, University of Toronto,}\\
\normalsize{6 King's College Rd, Toronto, Ontario M5S 3G4, Canada}\\
}

\date{}

\title{{\Huge Multi-task Neural Networks for QSAR Predictions}}

\begin{document}

\maketitle

\begin{abstract}
  Although artificial neural networks have occasionally been used for
  Quantitative Structure-Activity/Property Relationship
  (QSAR/QSPR)studies in the past, the literature has of late been
  dominated by other machine learning techniques such as random
  forests. However, a variety of new neural net techniques along with
  successful applications in other domains have renewed interest in
  network approaches. In this work, inspired by the winning team's use
  of neural networks in a recent QSAR competition, we used an
  artificial neural network to learn a function that predicts
  activities of compounds for multiple assays at the same time. We
  conducted experiments leveraging recent methods for dealing with
  overfitting in neural networks as well as other tricks from the
  neural networks literature. We compared our methods to alternative
  methods reported to perform well on these tasks and found that our
  neural net methods provided superior performance.
\end{abstract}

\section{Introduction}
 % intro.tex

Quantitative Structure Analysis/Prediction Studies attempt to build
mathematical models relating physical and chemical properties of
compounds to their chemical structure. Such mathematical models 
could be used to inform pharmacological studies by providing an
\emph{in silico} metholodogy to test or rank new compounds for desired 
properties without actual wet lab experiments. Already QSAR studies 
are used extenstively to predict pharmacokinetic properties such as ADME 
(absorption, distribution, metabolism and excretion) and toxicity. 
Improvements in these methods could provide numerous benefits to the
drug development pipeline. 

Learning QSAR models requires three main steps: generating a training
set of measured properties of known compounds, encoding information
about the chemical structure of the compounds, and buildilng a
mathematical model to predict measured properties from the encoded
chemical structure. High throughput screening (HTS) studies are ideal
for collecting training data --- a few hundred to thousands of
compounds are now tested routinely using HTS equipment against an
assay of interest (which may be cellular or biochemical). Generating
molecular descriptors from compound structures is also a well studied
problem and various methods of encoding relevant information about
compounds as vectors of numbers have been developed. These descriptors
measure various physical, chemical, and toplogical properties of the
compounds of interest. Once descriptors have been computed, machine
learning and statistics supply the mathematical models used to make
the predictions. The particular machine learning techniques used
matter a great deal for the quality of QSAR predictions. In a QSAR
competition sponsored by
Merck\footnote{\url{http://www.kaggle.com/c/MerckActivity} (Retrieved
  \today)}, all competitors used the same descriptors and training
data, but nevertheless our machine learning techniques (in particular
deep neural network models related to the neural nets we describe in
this paper) allowed the winning
team\footnote{\url{http://blog.kaggle.com/2012/11/01/deep-learning-how-i-did-it-merck-1st-place-interview/}
  (Retrieved \today)} to achieve a relative accuracy improvement of
approximately 15\% over Merck's internal baseline.

Applying machine learning algorithms to QSAR has a rich history.
Early work applying machine learning to QSAR tasks used linear
regression models, but these were quickly supplanted by Bayesian
neural networks \citep{bayesiannn,brann,bayesiannn_ard} and other approaches. Recently,
random forests \citep{RandomForestQSAR2003}, projection pursuit, partial least
squares and support vector machines \citep{lssvm1,lssvm2} have also been
applied successfully to this task. Each of these methods has different
advantages and disadvantages (see \citet{review_qsar_methods}
for a recent review). Practitioners in the field have often been partial to methods
that, in addition to making accurate predictions, allow for variable
selection, so that users are able to assess whether chosen features
are indeed useful from the viewpoint of an informed chemist, and to
methods that allow for easy assessment of uncertainty from the
models. Also important has been the issue of controlling overfitting,
although this is a concern that is paramount not only to QSAR, but all
domains where high-capacity machine learning models are applied to
small data sets.  Random forests have, thus far, been well-suited to
these requirements since they do not readily overfit and since they
provide a simple measure of variable importance.  Bayesian neural
networks are particulary suitable for assessing uncertainty about
model predictions and controlling overfitting.  However, the
computational demands of Bayesian neural networks necessitate small
hidden layers and using variable selection to reduce the input
dimensionality. Gaussian process regression models can be viewed as an
infinite hidden layer limit of Bayesian neural networks, but can still
be quite computationally expensive, often requiring time cubic in the
number of training cases \citep{gpr_qsar}.

Non-Bayesian neural networks have also been applied to QSAR
\citep{devillers1996neural}, but often had only a single small hidden layer
(presumably to avoid overfitting issues that Bayesian versions
obviate).  The past decade has seen tremendous improvements in
training methods for deep and wide neural networks as well as a
renewed appreciation for the advantages of deeper networks. Deep
neural networks (DNNs) have been highly successful on a variety of
different tasks including speech recognition \citep{DeepSpeechReview},
computer vision \citep{KrizhevskyEtAl2012}, and natural language processing
\citep{CollobertWeston2008ICML}. The methodological
improvements, along with faster machines, have allowed researchers to
train much larger and more powerful models --- instead of neural
networks with only one hidden layer with 20-50 hidden units, neural
networks are now regularly trained with many hidden layers of
thousands of hidden units, resulting in millions of tunable
parameters. Networks this large can even be trained on small datasets
without substantial overfitting using early stopping, weight
penalties, and the newer techniques of unsupervised
pre-training \citep{Hinton06} and dropout \citep{dropout}. All of these
techniques for avoiding overfitting in large neural nets trained on
small datasets are relatively inexpensive computationally, especially
compared to a fully Bayesian approach.

As mentioned above, the neural net QSAR models we trained in this work
differ from those used in the QSAR literature in a variety of
respects. However, the most important difference is that the neural
nets we explore in this paper are multi-task neural networks that
operate on multiple assays simultaneously, something that is very rare
in the literature. The only other work we are aware of to use
multi-task neural nets for QSAR is \citet{ChemInfModel2006} and there
are a variety of differences between it and the work we present
here. In particular, we used the recently developed dropout technique
to control overfitting and we did not have access to target protein
features. Additionally, \citet{ChemInfModel2006} focussed on exploring
the performance of the kernel-based JRank algorithm for collaborative
filtering and not different neural net variants.

The motivation behind multi-task learning is that
performance on related tasks can be improved by learning shared
models. Multi-task neural network models are particularly
appealing because there can be a shared, learned feature extraction
pipeline for multiple tasks. Not only can learning more general
features produce better models, but weights in multi-task nets are
also constrained by more data cases, sharing statistical
strength. QSAR tasks naturally lend themselves to multi-tasking ---
assays are often related to each other and different compounds might
share features. Even if assays are only tenuously related, they are
still governed by the laws of chemistry and it might still be important
to learn more broadly useful higher-level features from the initial
descriptors.

In this paper, we apply multi-task learning to QSAR using 
various neural network models.  We do this while leveraging some of the recent 
developments outlined above.  Our results show that 
neural nets with multi-tasking can lead to significantly improved
results over baselines generated with random forests.

\section{Methods}
\label{sec:methods}
 % methods.tex

The machine learning model that we use as the basis for our approach
to QSAR is the feedforward (artificial) neural network (ANN). Neural
networks are powerful non-linear models for classification,
regression, or dimensionality reduction. A neural network maps input
vectors to output vectors with repeated compositions of simpler
modules called layers. The internal layers re-represent the input and
learn features of the input useful for the task. 

Mathematically, an $L$-layer neural network is a vector valued
function of input vectors $\vec{x}$, parameterized by weight matrices
$\mathbf{W}_l$ and bias vectors $\vec{b}_l$, defined by:
\begin{align*}
\vec{y}_0 &= \vec{x} \\
\vec{z}_l &= \mathbf{W}_l \vec{y}_{l-1} + \vec{b}_l \\
\vec{y}_{l} &= h_l ( \vec{z_l} ),
\end{align*}
where the $h_l$ are elementwise, nonlinear \emph{activation functions}
(such as $\frac{1}{1+e^{-x}}$), $\vec{y}_{l}$ is the \emph{activation}
of layer $l$ and $\vec{z}_l$ is the \emph{net input} to layer $l$.
Given a dataset of known input and output vectors $\left\{ (\vec{x}_n
  , \vec{t}_n) \right\}$, we optimize the parameters $\Theta = (\{
\mathbf{W}_l \}, \{ \vec{b}_l \})$ of the network to minimize some
cost function. For standard regression problems, the output
$\vec{y}(\vec{x}) = \vec{y}_{L}(\vec{x}_n; \Theta)$ and target vectors
will be one dimensional. A typical cost function for regression is the
mean squared error:
\[ C_{MSE}(\Theta) = \frac{1}{2} \sum_{n=1}^{N} \| \vec{y}_{L}(\vec{x}_n; \Theta) - \vec{t}_n \|^2 .\]
For classification problems, the cross-entropy error function is more
appropriate. In the binary classification case, the cross entropy
error is:
\[ C_{CE}(\Theta) = -\sum_{n=1}^{N} t_n \log y(\vec{x}_n; \Theta) + (1 -
t_n) \log (1-y(\vec{x}_n; \Theta)) .\] We trained all neural networks
in this work using minibatched stochastic gradient descent (SGD) with momentum. In
other words, we repeatedly estimated $\frac{\partial
  C(\Theta)}{\partial \Theta}$ from a small minibatch of training
cases and used it to update the ``velocity'' of the parameters (see
the appendix A for details on the update rule).

\subsection{Multi-task neural networks}
The most direct application of neural networks to QSAR modeling is to
train a neural net on data from a single assay using vectors of
molecular descriptors as input and recorded activities as training
labels. This single-task neural net approach, although simple, depends
on having sufficient training data in a single assay to fit the model
well. We suspect data scarcity is one reason agressively regularized
models such as random forests are popular in the
literature \citep{RandomForestQSAR2003}. In order to leverage data from
multiple assays, we use a multi-task neural net QSAR architecture that
makes predictions for multiple assays simultaneously. Vectors of
descriptors still serve as the input to the network, but there is a
separate output unit for each assay. For any given compound, we
typically only observe one or at most a small number of output values
since in general we do not expect compounds to appear in all assays
regularly.  During multi-task neural net training we only perform
backpropagation through output layer weights incident on output units
whose value we observe for the current training case. For simplicity
we treat a compound that appears in $k$ different assays as $k$
different training cases that have the same descriptor vector but
different outputs observed. Since the number of compounds screened
varies across assays, na\"ively combining all the training cases from
all the available assays and training a multi-task neural net would
bias the objective function towards whatever assay had the most
compounds. We handle this issue by controlling how many training cases
from each assay go into a minibatch. For example, if there are 7
assays we could create minibatches of 80 cases by drawing 20 cases at
random (with replacement) from a particular assay we wish to emphasize
and 10 cases from each of the other six assays.

\citet{profileQSARMartin2011} presented another way of leveraging
related assays called Profile-QSAR that is similar in spirit to the
multi-task neural net approach we use (although it does not use neural
nets), but has many important differences. Profile-QSAR treats some of
the assays as side information and uses single task methods to
complete an assay/compound activity matrix for these previously
studied assays. The imputed and measured activities for a particular
compound in the side information assays become additional descriptors
to use when making predictions for the compound. Unlike Profile-QSAR,
the multi-task neural net approach does not do any imputation of
activities. Another difference is that a multi-task neural net trains
on all available assays and potentially makes predictions on all
available assays as well while learning a shared feature extraction
pipeline.

\subsection{Wider and deeper neural networks}
Deep neural networks, or neural networks with multiple hidden layers,
have been highly successful recently in numerous applications (notably
computer vision \citep{KrizhevskyEtAl2012} and speech recognition \citep{DeepSpeechReview})
because they are capable of learning complicated, rapidly-varying
non-linear functions and are also capable of extracting a hierarchy of
useful features from their input. Many successful deep neural network
models have millions of tunable parameters and wide layers with
thousands of units. In contrast, neural nets used in QSAR to date tend
to have only a single layer and sometimes very few hidden units
(for example \citet{small_nn}). Recent advances in 
training and regularizing wide and
deep neural nets and in computer hardware have changed the dominant
approach to training neural networks in the machine learning community
from training small nets that are incapable of overfitting and often
underfit to instead agressively regularizing wide and deep neural
networks with many tunable parameters. The most important lesson from
the recent successes of deep learning is that, although deeper
networks will not always perform better than shallow ones,
practitioners need to be using models that can trade breadth for depth
and vice versa, since one particular architecture cannot be the best
for all problems. In this paper, we describe our initial attempts to
leverage advances in deep neural network methods in QSAR applications.
As assays become cheaper and more assay results accumulate, machine
learning models with high capacity will have the best performance.
Many of the models we trained (including nets with only a single
hidden layer) have far more weights than we have training cases, but
through careful training and regularization they can still perform
well.

\subsection{Regularizing large neural networks}

Using wide and/or deep neural nets with many tunable parameters makes
avoiding overfitting especially crucial. In QSAR modeling we often
want to use large and expressive models capable of representing
complex dependencies between descriptors and activities, but data may
also be quite limited. Regularization, broadly construed, has been the
subject of much recent deep learning research. Generative unsupervised
pre-training \citep{Hinton06,Erhan-aistats-2010} is a powerful
data-dependent regularizer that brought a lot of attention to deep
neural net models in particular. Early stopping of training based on
validation error can partially control overfitting but has limited
effectiveness when there is very little validation data and the
networks being trained have very high capacity. Penalizing large
weights in the model can also help avoid overfitting, although more
sophisticated techniques are, in our experience, much more useful. One
such technique is dropout \citep{dropout}. Dropout randomly corrupts
the activations of neurons in the network during training by zeroing
out their activities independently. Intuitively, one effect the noise
added by dropout has is that it penalizes large weights that result in
uncertain predictions or hidden unit activations. Another way to view
dropout is as approximate model averaging over the exponentially
numerous different neural nets produced by deleting random subsets of
hidden units and inputs. We can also view multi-task neural nets as a
way of regularizing the weights that are shared across assays. Since
only the output weights are assay-specific, using data from other
assays is a powerful way of avoiding overfitting. Instead of shifting
the weights towards zero the way an L2 penalty on the weights would,
multi-task neural nets shift the weights of the hidden unit feature
detectors towards values that extract features useful for other QSAR
tasks. In this way, hidden layers in a multi-task neural network
in-effect learn higher level, more abstract molecular descriptors and
have a training objective that encourages them to create features
useful for more than one task.

\section{Experiments}
\label{sec:experiments}
 % experiments.tex
In order to test the effectiveness of multi-task neural nets for
QSAR we trained neural nets on publically available assay results and
compared their performance to a selection of baseline methods.

\subsection{Dataset Generation}
We ran experiments on assay results deposited in PubChem
(\url{http://pubchem.ncbi.nlm.nih.gov/}).  We used 19 different assays,
selected so at least several of them were closely related. Table
\ref{tbl:assays} lists the assays that were used in our
experiments. We included both cellular and biochemical assays and in
some cases used multiple related assays, for example assays for
different families of cytochrome P450 enzymes.

\begin{table}
\scriptsize
  \caption{List of assays from Pubchem that were used for this study}
  \label{tbl:assays}
  \begin{tabular}{lllll}
    \hline
    AID       & Assay Target / Goal & Assay Type & \# active & \# inactive\\
    \hline
    1851(2c19)& Cytochrome P450, family 2, subfamily C, polypeptide 19     & Biochemical  & 5913 & 7532 \\
    1851(2d6) & Cytochrome P450, family 2, subfamily D, polypeptide 6,& Biochemical & 2771 & 11139\\
                     & isoform 2 \\
    1851(3a4) & Cytochrome P450, family 3, subfamily A, polypeptide 4      & Biochemical  & 5266 & 7751 \\
    1851(1a2) & Cytochrome P450, family 1, subfamily A, polypeptide 2      & Biochemical  & 6000 & 7256 \\
    1851(2c9) & Cytochrome P450, family 2, subfamily C, polypeptide 9      & Biochemical  & 4119 & 8782\\
    1915      & Group A Streptokinase Expression Inhibition                & Cell         & 2219 & 1017\\
    2358      & Protein phosphatase 1, catalytic subunit, alpha isoform 3  & Biochemical  & 1006 & 934\\
    463213    & Identify small molecule inhibitors of tim10-1 yeast        & Cell         & 4141 & 3235 \\
    463215    & Identify small molecule inhibitors of tim10 yeast          & Cell         & 2941 & 1695\\
    488912    & Identify inhibitors of Sentrin-specific protease 8 (SENP8) & Biochemical  & 2491 & 3705\\
    488915    & Identify inhibitors of Sentrin-specific protease 6 (SENP6) & Biochemical  & 3568 & 2628\\
    488917    & Identify inhibitors of Sentrin-specific protease 7 (SENP7) & Biochemical  & 4283 & 1913\\
    488918    & Identify inhibitors of Sentrin-specific proteases (SENPs)  & Biochemical  & 3691 & 2505\\
              & using a Caspase-3 Selectivity assay\\
    492992    & Idenfity inhibitors of the two-pore domain potassium &Cell & 2094 & 2820\\
    & channel (KCNK9) \\
    504607    & Identify inhibitors of Mdm2/MdmX interaction               & Cell    & 4830 & 1412\\
    624504    & Inhibitor hits of the mitochondrial permeability & Cell & 3944 & 1090 \\
                    & transition pore \\
    651739    & Inihibition of Trypanosoma cruzi & Cell & 4051 & 1324\\
    651744    & NIH/3T3 (mouse embryonic fibroblast) toxicity & Cell & 3102 & 2306\\
    652065    & Identify molecules that bind r(CAG) RNA repeats & Cell & 2966 & 1287\\
    \hline
  \end{tabular}
\end{table}

We generated molecular descriptors with Dragon\footnote{\url{http://www.talete.mi.it/}}
to use as input to the various machine learning models.  While Dragon
can generate 4885 different descriptors, several of these descriptors
were inapplicable to all of the compounds in our data sets. After
excluding the inapplicable ones, we were left with 3764 molecular
descriptors. Each descriptor was Z-score normalized over all the
compounds in the union of all the assays. For some of the single task
neural net models, we also generated additional binary features
by thresholding other single descriptors. We selected descriptors and
thresholds from decision nodes used in a boosted ensemble of limited
depth decision trees.

\subsection{QSAR Model Training}
Using the assays, we treated
the problem as a classification task using the active/inactive labels
produced by the assay depositors. QSAR prediction can be formulated as
a classification problem, a regression problem, or a ranking problem.
All of our techniques are equally applicable to any of these problem
formulations, but in this paper we only consider the binary
classification version of the problem. A natural way to use a model
that predicted activities would be to use it for virtual screening
which is ultimately a ranking problem.  Although the models we trained
optimized classification performance, we use area under the ROC curve
(AUC) as a performance metric since it emphasizes the ranking aspect
of the problem more relevant to virtual screening applications.

For each assay, we held out at random 25\% of the ligands to use as a
test set, leaving the remaining 75\% as a training set. We used
several classifiers implemented in the scikits-learn
\citep{scikit-learn} package as baselines: random forests (RF),
gradient boosted decision tree ensembles (GBM), and logistic
regression (LR). We split the training set into four folds and trained
each model four times with a different fold held out as validation
data. We average the test set AUCs of the four models when reporting
test set results. We used performance on the validation data to select
the best particular model in each family of models.  To the extent
that the baseline models required metaparameter tuning (e.g. selecting
the number of trees in the ensemble), we performed that tuning by hand
using validation performance. 

\subsubsection{Neural Network Metaparameter Selection}
Neural networks can have many important metaparameters, including
architectural metaparameters, such as layer sizes and hidden unit link
functions, optimization metaparameters, such as learning rates and
momentum values, and regularization metaparameters such as the dropout
probabilities for each layer, how long to train before stopping, and
learning rate annealing schedules. We trained neural nets that used
rectified linear units as well as neural nets that used sigmoidal
units and most neural nets had between two and eight million
parameters. In order to have an experimental protocol that avoids as
many of the viccissitudes of human expert judgement as possible, we
set all neural net metaparameters using Bayesian optimization
\citep{snoek-etal-2012b-spearmint,snoek-etal-2013b-spearmintWarp} of
the validation AUC.  Bayesian optimization is a type of sequential,
model-based optimization ideally suited for globally optimizing
blackbox, noisy functions while being parsimonious in the number of
function evaluations. For metaparamter optimization, Bayesian
optimization constructs a model of the function mapping metaparameter
settings to validation AUC and suggests new jobs to run based on an
acquisition function that balances exploration of areas of the space
where the model is highly uncertainty with areas of the space likely
to produce better results than the best job run so far. We used the
Spearmint software that implements the Bayesian optimization
algorithms from \citet{snoek-etal-2012b-spearmint} and in particular
we use the constrained version with warping enabled. See the appendix
B for details of what neural net metaparameters we optimized over what
ranges.

Bayesian optimization finds good model configurations very
efficiently, but makes very strong use of the validation error and,
with small validation sets, can quickly overfit the validation
data. However, overfitting the validation data will only hurt the test
set performance of the neural net models relative to the baselines
that have fewer metaparameters that can be tuned readily and
repeatably by hand.

\section{Results and discussion}
\label{sec:results}
 % resultsDisc.tex

Table \ref{tab:mainResults} contains our main results on all 19 assays
we investigated. On 14 of the 19 assays, the best neural net achieves
a test set AUC exceeding the best baseline result by a statistically
signifigant margin. We measured statistical signifigance with
$t$-tests using standard errors from training the models repeatedly on
new bootstrap samples (see the appendix C for more
details). On the remaining five assays there was no statistically
signifigant difference between the best decesion tree ensemble
baseline and the best neural net. On all assays, logistic regression
was by far the worst performing model. On assays with closely related
assays available, multi-task neural nets often performed better than
their single-task counterparts, even with no a priori information on
which particular assays were related to each other and despite having
to make good predictions on all assays simultaneously, including
unrelated ones.

The three assays (assay ids 1915, 2358, and 652065) that the decision
tree baselines did best on relative to the neural net models (despite
the differences not being statistically signifigant) had some of the
fewest training cases and had no closely related assays in our
dataset. Since we generally expect high-capacity, multi-task neural
net models to provide benefits for larger datasets and when there are
related assays to leverage, the three assays the baselines do
comparitively better on are not surprising.

\begin{table}%[h]
\centering
\begin{tabular}{|l|r|r|r|r|}
\hline
Assay         & RF    & GBM   & NNET  & MULTI \\ \hline
1851\_1a2  & 0.915 & 0.926 & 0.926 & \bftab 0.938 \\ \hline
1851\_2c19 & 0.882 & 0.894 & 0.897 & \bftab 0.903 \\ \hline
1851\_2c9  & 0.876 & 0.891 & 0.889 & \bftab 0.907 \\ \hline
1851\_2d6  & 0.839 & 0.857 & \bftab 0.863 & 0.861 \\ \hline
1851\_3a4  & 0.871 & 0.896 & 0.895 & 0.897 \\ \hline
1915       & 0.754 & 0.757 & 0.756 & 0.752 \\ \hline
2358       & 0.745 & 0.764 & 0.738 & 0.751 \\ \hline
463213     & 0.651 & 0.659 & 0.651 & \bftab 0.676 \\ \hline
463215     & 0.614 & 0.610 & 0.613 & \bftab 0.654 \\ \hline
488912     & 0.664 & 0.672 & 0.664 & \bftab 0.816 \\ \hline
488915     & 0.700 & 0.713 & 0.723 & \bftab 0.873 \\ \hline
488917     & 0.818 & 0.834 & 0.835 & \bftab 0.894 \\ \hline
488918     & 0.785 & 0.800 & 0.784 & \bftab 0.842 \\ \hline
492992     & 0.804 & 0.827 & 0.803 & 0.829 \\ \hline
504607     & 0.673 & 0.670 & \bftab 0.684 & 0.670 \\ \hline
624504     & 0.851 & 0.869 & 0.871 & \bftab 0.889 \\ \hline
651739     & 0.775 & 0.793 & 0.790 & \bftab 0.825 \\ \hline
651744     & 0.870 & 0.885 & 0.886 & \bftab 0.900 \\ \hline
652065     & 0.787 & 0.793 & 0.793 & 0.792 \\ \hline
\end{tabular}
\caption{This table shows the average test set AUC for random forests
  (RF), gradient boosted decision tree ensembles (GBM), single-task
  neural nets (NNET), and multi-task neural nets (MULTI) on
  each assay. The multi-task neural nets are trained to make
  predictions for all assays at once. If the best neural net result on
  a particular assay is better than both decision tree baselines and
  the differences are statistically significant then we use boldface
  (the difference between the two nets may or may not be statistically
  significant). In the cases where the best decision tree baseline is
  better than both neural nets the differences are not statistically
  significant.}
\label{tab:mainResults}
\end{table}

\subsection{Multi-tasking vs combining assays}
The 19 assays we used contain several groups of closely related
assays. Specifically, we have five cytochrome P450 assays with minor
variations, four assays for inhibition of Sentrin-specific proteases,
and two related cellular assays for inhibition of particular yeast
strains. When two assays are strongly \emph{positively} correlated
(which will happen when they are nearly identical and share enough
compounds), simply merging the data from both might work as well as
the more sophisticated multi-task neural net method that can leverage
more general positive, negative, and feature-conditional
correlations. In fact, if all the related assays we worked with were
nearly identical, the gains we see from the multi-task neural net over
the single-task methods might simply reflect the gains expected from
adding more data to a single-task classifier. To investigate this
issue, we created additional datasets that combined data from multiple
assays. For example, the 488912 assay has three related assays:
488915, 488917, and 488918. We added the training data from 488915,
488917, and 488918 to the 488912 training set while still testing on
data exclusively from 488912. In this manner, we created 11 combined
datasets and trained GBMs on them, since GBMs were the best performing
non-neural net model.

Table \ref{tab:multiVsCombined} shows the results on these 11 datasets
for single-task GBMs, multi-task neural nets using all 19 assays, and
GBMs combining training data only from related assays. On 8 of the 11
datasets, the multi-task neural nets exhibit a statistically
signifigant improvement over the single-task GBM and the combined
training set GBM. On the remaining three datasets, there is not a
statistically signifigant difference in performance between the best
of the three models and the second best. Table
\ref{tab:singleVsMultiNNet} highlights results (also displayed in
table~\ref{tab:mainResults}) comparing single- and multi-task neural
nets on assays with other related assays in our collection. On all but
two of the 11 assays, the multi-task neural net trained using all 19
assays obtain statistically signifigant improvements in test AUC over
the single-task neural nets. Since the multi-task neural net models
can learn to ignore the other assays when making predictions for a
particular assay, at worst they will only get somewhat weaker results
than a single-task neural net because they waste capacity on
irrelevant tasks. And, indeed, that is what we see in our results.

The 48891* series of assays are closely related enough for the GBMs
trained on the combined training set of the four assays to do much
better than the GBMs that use only the primary assay's training set.
On some of the 48891* series assays, a GBM trained on the combined
training sets even does better than a multi-task neural net trained to
make predictions on all 19 datasets, although the improvement is not
statistically signifigant. However, the multi-task neural net, in
addition to having to make predictions for other unrelated assays, is
not told that all of the 48891* series assays should be treated as the
same problem. In contrast to the 48891* series, on the 1851\_* series
of assays, the GBMs trained on the combined 1851\_* training sets do
worse than the GBMs that just use the primary training set. On these
assays, however, the multi-task neural net improves upon on both the single
task GBM and neural net, showing that it can leverage information from
the related assays in a more nuanced way. In practice we will often
have somewhat related assays that are nevertheless far from identical
and in this situation the multi-task neural net models can stil
provide benefits, unlike a classifier simply combining the training
sets from different assays.

On the 46321* series of assays, although the GBMs on the combined
training sets were better than the GBMs trained only on data from the
primary assay, the multi-task neural net was better still. This result
demonstrates that the two assays in the group are not sufficiently
contradictory to confuse the GBM models on the combined training set
(as happened with the 1851\_* series), but there are still gains to be
had from the multi-task neural net. The most interesting cases arise
when assays are related but not positively correlated strongly enough
to simply combine into a single dataset. In this scenario, multi-task
neural nets shine because they can use negative and partial
correlations.

\begin{table}%[h]
\centering
\begin{tabular}{|l|r|r|r|}
\hline
Primary Assay & GBM   & MULTI & GBM Combined \\ \hline
1851\_1a2     & 0.926 & \bftab 0.938 & 0.905        \\ \hline
1851\_2c19    & 0.894 & \bftab 0.903 & 0.883        \\ \hline
1851\_2c9     & 0.891 & \bftab 0.907 & 0.879        \\ \hline
1851\_2d6     & 0.857 & \bftab 0.861 & 0.840        \\ \hline
1851\_3a4     & 0.896 & 0.897 & 0.869        \\ \hline
463213        & 0.659 & \bftab 0.676 & 0.665        \\ \hline
463215        & 0.610 & 0.649 & 0.648        \\ \hline
488912        & 0.672 & \bftab 0.825 & 0.815        \\ \hline
488915        & 0.713 & 0.873 & 0.868        \\ \hline
488917        & 0.834 & \bftab 0.915 & 0.909        \\ \hline
488918        & 0.800 & \bftab 0.869 & 0.852        \\ \hline
\end{tabular}
\caption{Multi-task neural nets compare favorably to GBMs using training sets that combine related assays. Bold entries correspond to statistically signifigant differences.}
\label{tab:multiVsCombined}
\end{table}

\begin{table}%[h]
\centering
\begin{tabular}{|l|r|r|}
\hline
Primary Assay & NNET  & MULTI \\ \hline
1851\_1a2     & 0.926 & \bftab 0.938 \\ \hline
1851\_2c19    & 0.897 & \bftab 0.903 \\ \hline
1851\_2c9     & 0.889 & \bftab 0.907 \\ \hline
1851\_2d6     & 0.863 & 0.861 \\ \hline
1851\_3a4     & 0.895 & 0.897 \\ \hline
463213        & 0.651 & \bftab 0.676 \\ \hline
463215        & 0.613 & \bftab 0.649 \\ \hline
488912        & 0.664 & \bftab 0.825 \\ \hline
488915        & 0.723 & \bftab 0.873 \\ \hline
488917        & 0.835 & \bftab 0.915 \\ \hline
488918        & 0.784 & \bftab 0.869 \\ \hline
\end{tabular}
\caption{For assays related to other assays in our collection, multi-task neural nets typically provide statistically signifigant improvements over single-task neural nets. Bold entries correspond to statistically signifigand differences.}
\label{tab:singleVsMultiNNet}
\end{table}

\subsection{Controlling overfitting without feature selection}

Since we allowed Bayesian optimization to train fully connected neural
networks with as many as about 3500 hidden units in a single layer, we
used a variety of methods to prevent overfitting. Bayesian
optimization quickly learned to enable dropout and use a non-zero L2
weight penalty. For any given assay, the best performing neural net
always used some dropout and in preliminary hand-tuned experiments
dropout seemed crucial as well.

\begin{figure}
\begin{center}
\includegraphics[width=0.84\textwidth]{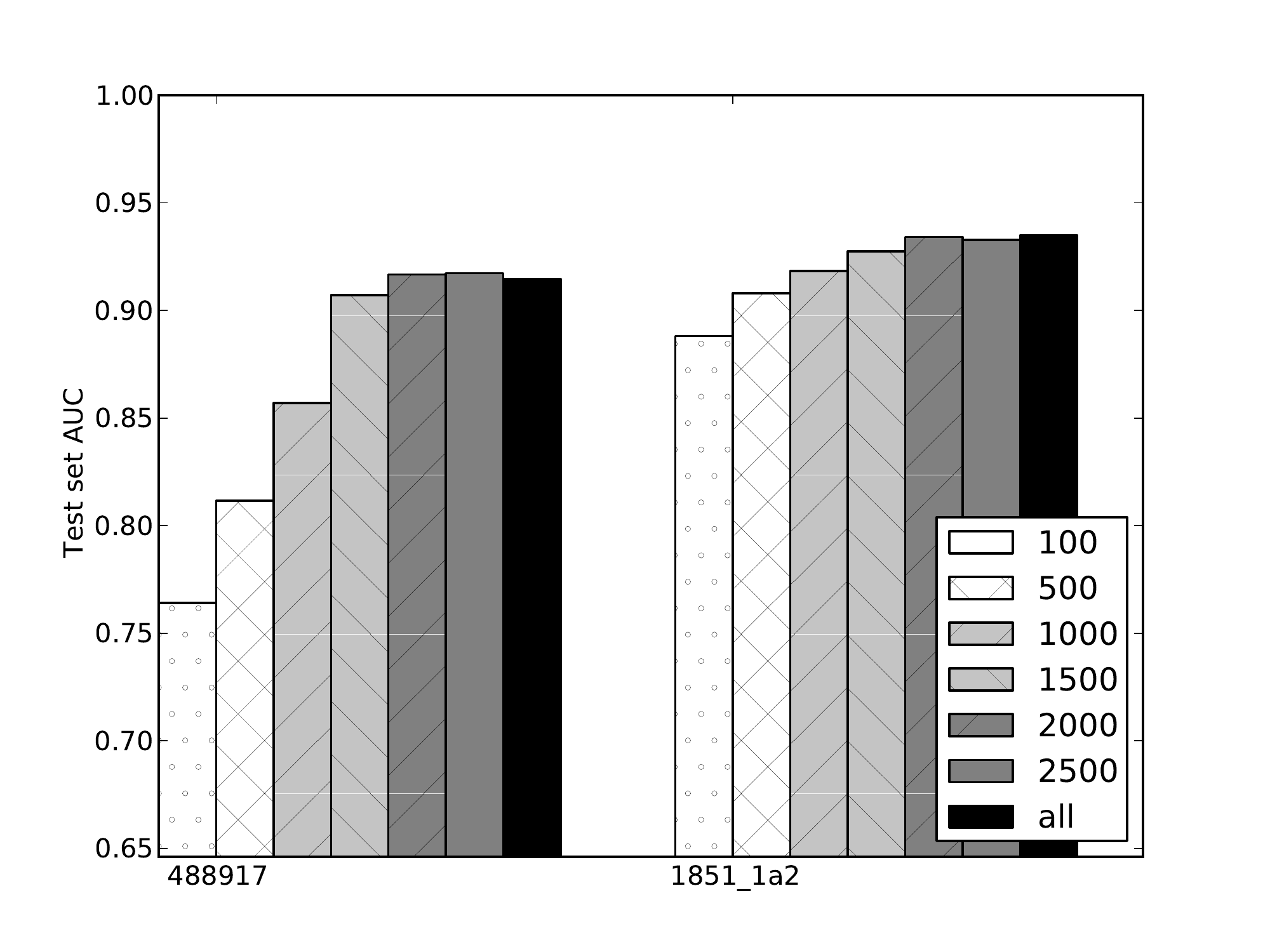}
\end{center}
\caption{Test AUC on two representative assays of a multi-task neural
  net using different numbers of input features. For a given number of
  input descriptors, we selected the best features as measured by information gain.}
\label{fig:featSelect}
\end{figure}

Unlike a lot of QSAR work in the literature, for example
\citet{winkler2002role} who warns against including too many
descriptors even if they contain relevant information, we do not
advise performing feature selection to reduce the number of input
dimensions drastically. Although not all the descriptors are necessary
or very informative, well-regularized and properly trained neural
networks can handle thousands of correlated input features. We trained
neural nets using all (3764) descriptors as well as ones using the
2500, 2000, 1500, 1000, 500, or 100 most informative input features
(as measured by information gain). On most assays, using the 2000-2500
most informative input descriptors did not degrage test set AUC very
much, but using only 1500 or fewer typically produced a large an
unnecessary drop in test set AUC. Figure \ref{fig:featSelect} shows
the test AUC for two representative assays. We generated the plot by
training the best multi-task neural net using the relevant primary
assay on different numbers of input descriptors.

\subsection{Neural network depth}

For the 19 assays we used and the descriptor set we used, changing the
number of hidden layers had no consistent effect. We performed
separate Bayesian optimization runs with one, two, and three hidden
layer multi-task neural networks and one and two hidden layer single
task neural networks. For single task neural nets, adding a second
hidden layer had very little effect and the best result from each
Bayesian optimization run for a particular model class achieved about
the same test AUC regardless of the number of hidden layers. However,
although there was no consistent trend for multi-task nets, allowing
deeper models seemed to be occasionally important. Table
\ref{tab:depthMulti} shows the results for the multi-task nets of
different depths. Although the optimal depth is not the same or
predictable across assays, on some assays there are large differences
in performance between multi-task neural nets of different depths.

\begin{table}[h]
\centering
\begin{tabular}{|l|r|r|r|}
\hline
Assay      & \multicolumn{1}{|l}{1 hidden layer} & \multicolumn{1}{|l}{2 hidden layers} & \multicolumn{1}{|l|}{3 hidden layers} \\ \hline
1851\_1a2  & 0.932                               & \bftab 0.938                                & 0.930                                 \\ \hline
1851\_2c19 & \bftab 0.903                               & 0.884                                & 0.881                                 \\ \hline
1851\_2c9  & \bftab 0.907                               & 0.894                                & 0.905                                 \\ \hline
1851\_2d6  & 0.855                               & \bftab 0.861                                & 0.768                                 \\ \hline
1851\_3a4  & 0.897                               & 0.897                                & 0.821                                 \\ \hline
1915       & 0.752                               & 0.754                                & 0.740                                 \\ \hline
2358       & 0.751                               & 0.758                                & 0.748                                 \\ \hline
463213     & 0.676                               & 0.678                                & 0.672                                 \\ \hline
463215     & 0.654                               & 0.637                                & 0.649                                 \\ \hline
488912     & 0.768                               & 0.816                                & 0.825                                 \\ \hline
488915     & 0.823                               & \bftab 0.873                                & 0.828                                 \\ \hline
488917     & 0.886                               & 0.894                                & \bftab 0.917                                 \\ \hline
488918     & 0.828                               & 0.842                                & \bftab 0.869                                 \\ \hline
492992     & 0.829                               & 0.827                                & 0.826                                 \\ \hline
504607     & 0.670                               & 0.665                                & 0.616                                 \\ \hline
624504     & 0.889                               & 0.881                                & 0.893                                 \\ \hline
651739     & 0.825                               & 0.821                                & 0.806                                 \\ \hline
651744     & 0.895                               & \bftab 0.900                                & 0.894                                 \\ \hline
652065     & 0.794                               & 0.792                                & 0.791                                 \\ \hline
\end{tabular}
\caption{Multi-task neural network results for neural nets with different numbers of hidden layers. We bolded the best result in a row when there was a statistically signifigant difference in test AUC between it and the second best entry in the row.}
\label{tab:depthMulti}
\end{table}

These depth results somewhat contradicts our experience in the Merck
molecular activity prediction contest where we found using neural
networks with more than one hidden layer to be crucial in almost all
cases. Unfortunately, since neither the contest data nor the
descriptors used in the contest are public or available for research
and additional experiments, we can only speculate about the cause of
this discrepancy. There were several assays in the contest with many
more compounds than the assays we used from PubChem and larger
datasets are more likely to provide enough training information to
usefully fit additional hidden layers. Larger datasets affecting the
optimal neural net depth is consistent with depth mattering more for
the multi-task nets we trained since they use data from all the
assays. The contest used the regression formulation of the task,
unlike our experiments in this work that used the binary
classification formulation, exaccerbating the difference in the number
of bits of information in the training labels. Since we did not focus
our efforts on trying many different descriptor sets, there may be
types of descriptors Dragon does not compute that did exist in the
contest data. A larger set of descriptors than the ones Dragon
computes might improve our results. For example, the open source RDKit
(\url{www.rdkit.org}) software provides Morgan fingerprint descriptors
that the current version of Dragon does not compute and other
commercial software packages such as MOE \citep{MOE} could have
descriptor sets that add important information also.

\section{Conclusions and future work}
Our results demonstrate that neural networks using the latest
techniques from the deep learning community can improve QSAR
prediction accuracies and, in particular, there is a very natural and
effective way of leveraging data from multiple assays when training a
neural network QSAR model. However, many more experiments need to be
done before we can settle on exactly the best way to solve QSAR
problems with neural networks. Treating the task as a binary
classification problem and using potentially unreliable
active/inactive decisions from assay depositors in PubChem may not be
the best way to approach the problem. In a virtual screening
application, a notion of how active a compound is towards a
particular target is essential and we plan to perform future work with
the ranking version of the problem. Given the effectiveness of
state-of-the-art Bayesian optimization software packages,
practitioners should no longer fear the large number of metaparameters
sophisticated neural network models can have since even with small
datasets we were able to find very good metaparameter settings
automatically. We also hope to develop better ways of implementing
multi-task neural nets that can make use of additional information
about which assays are likely to be related as well as target features
and other side information. Given the rapid progress of research on
neural network methods, we also hope to apply more advances from the
deep learning community to QSAR problems.

\subsection*{Acknowledgements}
  We would like to acknowledge Christopher Jordan-Squire and Geoff
  Hinton for their work on our team during the Merck molecular
  activity challenge; without the contest we would not have started
  work on this project.

\FloatBarrier

\bibliographystyle{plainnat}
\bibliography{qsar_dnn}

\appendix
\section{Stochastic gradient descent details}
For all neural net training, we used minibatch stochastic gradient
descent with momentum and backpropagation \citep{backprop} to compute
the necessary gradients. Let $C$ be the training objective function
and let $w$ be a generic neural net parameter. With $\langle
\frac{\partial C}{\partial w} \rangle$ denoting the average objective
function gradient over the current minibatch of cases, we used the
following weight update formulas:
\begin{align*}
v(t) &= \alpha v(t-1) + \epsilon \left( -\left\langle \frac{\partial C}{\partial
  w} \right\rangle - \lambda w \right) \\
w(t) &= w(t-1) + v(t),
\end{align*}
where $\epsilon$ is the learning rate or step size, $\alpha$ is the
momentum strength, and $\lambda$ is the weight cost strength.

\section{Bayesian optimization search space}

We used the constrained version of
Spearmint \citep{snoek-etal-2012b-spearmint} with warping enabled and
labeled training runs that diverged as constraint violations. We let
Spearmint optimize the metaparameters listed below with a budget of 30
(usually sequential) trials, although in a few preliminary experiments
we used a 50 trial budget. The allowed ranges were decided based on
our first single hidden layer, single-task neural net Spearmint run
and in some cases slightly adjusted for other Spearmint runs. In
general, since the adjustments were not major ones, it is safe to pick
the largest range we ever used for any given metaparameter; at worst a
few extra jobs would be required.

Metaparameters:
\begin{itemize}
\item dropout fractions $\in \left[ 0, 0.75\right]$, with a separate
  metaparameter for the input layer and each hidden layer
\item the number of training epochs in $\left[ 2, 100\right]$ for
  nets with a single hidden layer and in $\left[ 2, 120\right]$ for
  nets with two or more hidden layers
\item the number of hidden units in each layer (allowed to be
  different in different layers)

For single task neural nets, no hidden layer was allowed more than
3072 units. The minimum number of hidden units in a layer for a single
task neural net was 16 in our first single hidden layer run and 64 all
other times. For multi-task neural nets we had a minimum of 512 units
in each hidden layer and allowed up to 3584 units, except for the
three hidden layer models which we constrained to a maximum of 2048.

\item the annealing delay fraction $\in \left[ 0, 1 \right]$, or in
  other words the fraction of the training iterations that must
  complete before we start annealing the learning rate

We used a continuous parameterization of the fraction even though the
training program would end up rounding when computing what epoch to
start annealing the learning rate.

\item the initial learning rate  

The learning rate is applied to the average gradient over a
minibatch. We allowed initial learning rates in $\in \left[ 0.001,
  0.25 \right]$ for all multi-task neural net experiments and our
initial single-task, single hidden layer Spearmint run. All other
single-task neural net experiments allowed Spearmint to select an
initial learning rate in $\in \left[ 0.001,  0.3 \right]$.

\item the type of annealing, either exponential or linear

  The annealing mode was a discrete choice. For linear annealing, the
  plot of learning rate versus epoch is a straight, downward sloping
  line intersecting the initial learning rate when annealing starts
  and the final learning rate when training stops. Linear annealing
  used a final learning rate of $10^{-8}$. Once annealing has started,
  exponential annealing multiplies the learning rate by a constant
  shrinkage factor each iteration with the factor chosen to ensure
  that the learning rate when training stops is the final learning
  rate. Exponential annealing used a final learning rate of
  $10^{-6}$. We do not believe the precise value matters very much as
  long as it is small.

\item momentum $\in \left[ 0,  0.95 \right]$
\item the $L^2$ weight cost $\in \left[ 0,  0.005 \right]$ except for
 single-task neural nets with two hidden layers in which case we
 allowed weight costs in $\left[ 0,  0.007 \right]$

\item the hidden unit activation function, either logistic sigmoids or
  rectified linear units 

For simplicity, we forced all hidden units in a network to use the
same activation function.

\item scale (standard deviation) of the initial random weights $\in \left[ 0.01,  0.2 \right]$

We used Gaussian initial random weights. All weight matrices used the
same scale except for the weights from the inputs to the first hidden
layer which the subsequent metaparameter controls.
 
\item natural $\log$ of the multiplier for the scale of the bottom
  (input to hidden) weight matrix $\in \left[ -1,  1 \right]$

We allowed Spearmint to set the scale of the lowest weight matrix as
a multiplicative adjustment to the scale of the rest of the
weights. In this way, Bayesian optimization can set the bottom weight
matrix to have a scale up to $e$ times smaller or larger than the
other weight matrices.
\end{itemize}

\section{Statistical signifigance determination}
Since we presented tables of results for 19 different assays and the
assays have different numbers of data cases in them, it is important
to have at least some simple notion of what differences in test AUC
between models on a single assay are stronger or weaker evidence of a
true difference in performance. The simple bootstrap procedure
outlined below allowed us to indicated which differences between model
families were large enough to be potentially interesting in a way that
took into account the variability of our results due to the particular
training sample and the inherent variabilities of model training.

In order to get standard errors for the signifigance tests we
mentioned in the main text, we used bootstrap sampling. We trained on
8 different bootstrap samples of the complete training set. Let $y_1$
and $y_2$ be the average test set AUCs of two different models after
normal (not bootstrap samples) training on the different training
folds. Let $\sigma^2_1$ and $\sigma^2_2$ be the unbiased sample
variances corresponding to the test AUC results of the two models
after training on the new training sets that were sampled from the
original training set with replacement. We called a difference between
table entries $y_1$ and $y_2$ statistically signifigant when 
\[ | y_1 -y_2| > 1.96 \sqrt{\frac{\sigma^2_1 + \sigma^2_2}{8}}. \]
Although using the difference of cross validation mean test AUCs $y_1
-y_2$ on the left hand side in the test instead of the difference in
mean test AUC across the relevant bootstrap training runs is
non-standard, since the data sets we used are small, only training on
bootstrap samples noticeably degrades performance since a bootstrap
sample often will not contain all possible training cases. Our
approximate test rejects many seemingly large test AUC differences as
insignificant statistically on the assays with very little data (e.g.
2358 and 1915) which is reasonable behavior given the variability of
model test AUCs on datasets that are that small.

In our comparisons that involved GBMs using the combined training sets
from multiple assays, we did not recompute new standard errors and
instead used the single task GBM standard errors. Avoiding this
computation may overestimate the uncertainty in the test AUCs of the
GBMs using the combined training sets since training the same model
with more data generally reduces the training sample
variance. Overestimating the uncertainty in the test AUCs of the GBMs
trained on the combined data would only give them an
advantage when compared to the neural net models since the multi-task
neural nets compared to them had better results.

%\bibliography{qsar_dnn}

\end{document}